\documentclass[
]{ceurart}

\usepackage{listings}
\usepackage{multirow}

\begin{document}

\copyrightyear{2025}
\copyrightclause{Copyright for this paper by its authors.
  Use permitted under Creative Commons License Attribution 4.0
  International (CC BY 4.0).}

\conference{CLEF 2025 Working Notes, 9 -- 12 September 2025, Madrid, Spain}

\title{Team \textquote{better\_call\_claude}:  Style Change Detection using a Sequential Sentence Pair Classifier}

\title[mode=sub]{Notebook for the PAN Lab at CLEF 2025}

\tnotemark[1]
\tnotetext[1]{You can use this document as the template for preparing your
  publication. We recommend using the latest version of the ceurart style.}

\author[1]{Gleb Schmidt}[%
email=gleb.schmidt@ru.nl,
]
\cormark[1]
\fnmark[1]

\author[2]{Johannes Römisch}[%
email=johannes.roemisch@study.thws.de,
]
\fnmark[1]
\author[2]{Mariia Halchynska}[%
email=mariia.halchynska@study.thws.de,
]
\fnmark[1]

\author[3]{Svetlana Gorovaia}[%
email=sgorovaya@hse.ru,
]
\cormark[1]
\fnmark[1]

\author[2]{Ivan P. Yamshchikov}[%
email=ivan.yamshchikov@thws.de,
]

\address[1]{Humanities Lab, Faculaty of Arts, Radboud University, Houtlaan 4, 6525 XZ, Nijmegen, Netherlands}
\address[2]{Center for Artificial Intelligence, Technical University of Applied Sciences Würzburg-Schweinfurt,  Münzstraße 12, 97070, Würzburg, Germany}
\address[3]{LEYA Lab, School of Computer Science, Physics and Technology, HSE University, 6, 25th Liniya, Vasilievsky Ostrov, 199004, St Petersburg, Russia}

\cortext[1]{Corresponding author.}
\fntext[1]{These authors contributed equally.}

\begin{abstract}
Style change detection---identifying the points in a document where writing style shifts---remains one of the most important and challenging problems in computational authorship analysis. At PAN 2025, the shared task challenges participants to detect style switches at the most fine-grained level: individual sentences. The task spans three datasets, each designed with controlled and increasing thematic variety within documents. We propose to address this problem by modeling the content of each problem instance—that is, a series of sentences---as a whole, using a Sequential Sentence Pair Classifier (SSPC). The architecture leverages a pre-trained language model (PLM) to obtain representations of individual sentences, which are then fed into a bidirectional LSTM (BiLSTM) to contextualize them within the document. The BiLSTM-produced vectors of adjacent sentences are concatenated and passed to a multi-layer perceptron for prediction per adjacency. Building on the work of previous PAN participants classical text segmentation, the approach is relatively conservative and lightweight. Nevertheless, it proves effective in leveraging contextual information and addressing what is arguably the most challenging aspect of this year's shared task: the notorious problem of \textquote{stylistically shallow}, short sentences that are prevalent in the proposed benchmark data. Evaluated on the official PAN~$2025$ test datasets, the model achieves strong macro-F1 scores of $0.923$, $0.828$, and $0.724$ on the \texttt{EASY}, \texttt{MEDIUM}, and \texttt{HARD} data, respectively, outperforming not only the official random baselines but also a much more challenging one: \texttt{claude-3.7-sonnet}'s zero-shot performance.
\end{abstract}

\begin{keywords}
  Style Change Detection \sep
  Text Segmentation \sep
  Sequence Labeling \sep
  BiLSTM \sep
  Large Language Models \sep
  Pre-Trained Language Models \sep
\end{keywords}

\maketitle

\section{Introduction}

Detecting changes in writing style---in other words, identifying places within a document where stylistic signal changes---is a form of authorship analysis that, perhaps alongside authorship verification, holds the greatest potential for applications beyond industrial contexts, particularly in humanities research. Given that our contemporary concept of individual authorship---let alone formal definitions of intellectual property and copyright---is a relatively recent development, a substantial part of human written culture goes back to periods when, broadly speaking, \textquote{collaborative writing} (actual co-authorship, extensive reuse, interpolation to mention but a few of its possible forms), was not only common---as it remains today---but was often regarded as a way of declaring ones belonging to a tradition, and therefore valued even more highly than original composition.

Nonetheless, the exploration of such \textquote{mixed-authorship texts} is typically hindered precisely by the uncertainties surrounding their 
authorial structure, which creates notorious contextualization problems and subsequently puts strict limitations on the interpretation of such texts. 

Recent studies have shown that computational methods can offer valuable insights into mixed authorship at the level of entire corpora \cite{clerice_twenty-one_2023} or major subdivisions of individual works \cite{cafiero_psycheas_2021, plechac_relative_2021, schmidt_fine-tuning_2024}. However, this level of granularity may often be insufficient for solving research questions that require a more fine-grained diarization---at the paragraph or even sentence level \cite{eder_rolling_2016}.

In this context, the decade-long effort of PAN organizers to stimulate research in this direction deserves special recognition. In various forms, the style change detection task has consistently been a part of the competition's program since $2016$, making the participants' work notes and traditional overviews published in the aftermath of these events an invaluable source of methodological insight \cite{bevendorff_overview_2025, zangerle_multi-author_2025}. Echoing the field's growing theoretical sophistication, for almost a decade the PAN workshops have been offering increasingly complex benchmark data and task formulations, encouraging participants to push the boundaries of achievable.

\section{Related Work}
\subsection{Style Change Detection at PAN}
Since its first inclusion in the PAN program in $2016$, the style change detection task has appeared in various formulations. However, most of the factors contributing to its complexity were already present in the first two editions---namely, the required level of granularity for document analysis, the uncertainty regarding the number of style changes or contributing authors, and the need to segment the document. The most recent development of the task introduced only one additional---though important---dimension: controlled topic diversity in the data.

\subsubsection{2016—2022: In Search of Task's Score}
At PAN $2016$, the task was framed as an \emph{authorship diarization} problem, closely related to the traditional intrinsic plagiarism detection explored during the early years of the competition's history \cite{stamatatos_clustering_2016, rosso_overview_2016}. Participants faced three subtasks, each highlighting different challenges that were to become recurring focal points of the task in the following years. The first subtask assumed a major author and required identifying segments written by secondary authors. In the second subtask, the number of authors was provided, and participants were required to cluster document segments by authorship. The third and most challenging subtask involved building authorial clusters without any prior knowledge of the number of authors.

Operating at the sentence level, both proposed methods relied on traditional stylometric features, which were then processed using techniques typical of intrinsic plagiarism detection---namely, threshold- or Gaussian HMM-based outlier detection \cite{kuznetsov_methods_2016}, and clustering \cite{sittar_author_2016}. These approaches, however, failed to achieve sufficient performance at such a fine level of granularity.

The poor performance led to a redefinition of the task as a style breach detection problem at PAN $2017$. Instead of complete clustering a document’s segments by authorship, participants were asked to predict whether a document was written by multiple authors and, if so, to identify the points at which the writing style changes \cite{tschuggnall_overview_2017}.

The submitted approaches again centered on distance measures and outlier detection applied to sentences as well as actual or artificially constructed paragraphs, represented using either conventional stylometric features \cite{karas_opi-jsa_2017, khan_style_2017} or neural sentence embeddings \cite{safin_style_2017}. Despite the relative improvement over PAN $2016$, the results of PAN $2017$ confirmed that style-based document decomposition remained marginally beyond the state of the art at that time.

Therefore, for PAN $2018$, the task was redefined once again, framing the problem as a document-level binary classification task. This invited participants to explore how stylistic inconsistency could be detected across an entire text \cite{kestemont_overview_2018}. The submissions reflected both conventional feature engineering combined with rule-based or classical machine learning approaches  \cite{khan_model_2018, zlatkova_ensemble-rich_2018, safin_detecting_2018} and early applications of deep learning \cite{hosseinia_parallel_2018, schaetti_character-based_2018}. Ensembling multiple classifiers operating on diverse feature spaces---each capturing different aspects of language---not only proved reliable \cite{safin_detecting_2018, zlatkova_ensemble-rich_2018}, but also yielded the winning result \cite{zlatkova_ensemble-rich_2018}. The core of the classifier proposed by \cite{schaetti_character-based_2018} was a CNN designed to capture patterns of character bigrams in groups of varying length.

The submission by \cite{hosseinia_parallel_2018} deserves special attention not only because it scored second, but also because it anticipated developments in recent authorship analysis research and served as a distant source of inspiration for the approach proposed below. Instead of relying on traditional feature spaces---where lexical components took center stage at the time---\cite{hosseinia_parallel_2018} focused exclusively on the dependency trees of sentences. The expressiveness of this feature space has recently been confirmed in a series of authorship attribution studies \cite{gorman_author_2020, gorman_universal_2022, gorman_morphosyntactic_2024, gorman_morphosyntactic_2024-1}. \cite{hosseinia_parallel_2018} define and extract what they call a Parse Tree Feature (PTF)---a path from the root to any given word---and use it to represent each sentence as a sequence of its words’ PTFs, and each document as a sequence of sentence representations. Subsequently, both the original and reversed versions of the document are fed into an LSTM with an additional sentence-level attention mechanism, which contextualizes each sentence based on rich syntactic information across the entire problem. Finally, the similarity between the original and reversed representations is computed and used as the basis for prediction.

Responding to performance boost observed at PAN $2018$ on a simplified version of the problem, the organizers of PAN $2019$ increased the task's complexity once again by adding the objective of predicting the number of authors per document \cite{zangerle_overview_2019}. To address this task, \cite{nath_style_2019} employed a combination of clustering techniques based on representations of balanced-size text chunks obtained using the $50$ most frequent words (MFW).
Relying on a multi-layer perceptron operating on tf-idf-weighted word unigrams to detect style changes within a document, \citeauthor{zuo_style_2019} subsequently used an extension of \cite{zlatkova_ensemble-rich_2018}'s feature extraction procedure to represent document paragraphs. They then applied an ensemble model comprising two clustering methods and a multi-layer perceptron to predict the number of style changes. 

The Style Change Detection task at PAN $2020$ was marked by two important shifts. First, after a significant departure from its originally intended scope during PAN $2018$—$2019$, \textquote{the task was steered back into its original direction} \cite[1]{zangerle_overview_2020}. The segmentation component was reintroduced, and in addition to the document-level prediction of multi-authorship, participants were required to identify the positions where paragraph-level style changes occur. Second, for the first time, a solution based on pre-trained transformers was employed to address the task \cite{iyer_style_2020}, significantly outperforming clustering-based approaches---the $B_0$-maximal used by \cite{castro-castro_mixed_2020} and a modified version of \cite{nath_style_2019}, which, however, remained undocumented.

PAN $2021$ reintroduced yet another element of the task’s original scope that had previously been set aside as too complex: grouping of text segments by authorship within documents. The shared task presented arguably the most complete formulation of the problem, comprising three separate questions: (1) whether the text was written by multiple authors; (2) where between paragraphs the writing style changes; and (3) which author each paragraph belongs to \cite{zangerle_overview_2021}. Although the competition saw an increasing reliance on pre-trained transformers, it was marked by a wide diversity of methods. \cite{zhang_style_2021} proposed the highest-scoring approach for Tasks 2 and 3, using a similarity measure extracted from paragraphs with a pre-trained \texttt{BERT} model. They approached all tasks simultaneously, first solving Tasks 2 and 3 in an authorship verification fashion. Each paragraph was compared with all preceding ones, and a new authorial class was assigned whenever a paragraph could not be classified as written by the same author as any of the previous ones. This information was subsequently used to solve the remaining tasks. The approach by \cite{strom_multi-label_2021} excelled at Task 1. It combined sentence features extracted using \texttt{BERT} and aggregated per paragraphs with the set of stylometric features proposed by \cite{zlatkova_ensemble-rich_2018}. The tasks were solved by stacking two feature spaces and feeding them to an ensemble of four classifiers. \cite{singh_writing_2021} decomposed the tasks into a series of authorship verification problems and solved them adapting the method proposed by \cite{weerasinghe_feature_2020}. Other works operated over various selections of stylometric features and used LSTM-powered model \cite{deibel_style_2021} and Siamese architecture \cite{nath_style_2021} respectively. 

Three subtasks of PAN $2022$ challenged participants with both segmentation task and granularity level. Task 1 required finding the only style shift in a document co-authored by two persons. In Tasks 2 and 3, it was necessary to find style changes in a text written by two or more authors with switches occurring at only paragraph or paragraph and sentence levels. Despite clear prevalence of pre-trained transformer-based approaches, submissions exclusively working with manually-engineered feature spaces and traditional classification or clustering approaches \cite{alshamasi_ensemble-based_2022, alvi_style_2022} or combining hand-picked features with those extracted using pre-trained models were submitted \cite{rodriguez-losada_three_2022}. One of the submissions downright \textquote{hacked} the task by accessing extrinsic information online and yielding a nearly perfect result \cite{graner_unorthodox_2022}. Most submissions, however, explored different PLM-based architectures. The overall best score was achieved by \cite{lin_ensemble_2022} who obtained predictions for pairs of sentences or paragraphs by ensembling classifiers based on \texttt{BERT}, \texttt{RoBERTa}, and \texttt{ALBERT}. \cite{lao_style_2022} classified pairs of sentence or paragraph representations obtained applying one-dimensional convolution and max pooling to \texttt{BERT} output. \cite{zhang_style_2022} used a prompt-based approach fine-tuning a \texttt{BERT} model with masked language modeling objective to predict special tokens such \texttt{<DIFFERENT>} or \texttt{<SAME>} in a dynamically-constructed sequence: \textquote{They are the \texttt{<MASK>} writing style: {Para1} and {Para2}}. \cite{jiang_style_2022} trained three different transformer models to address each subtask. \cite{zia_style_2022} used LSTM, convolution, and max pooling over \texttt{BERT}-based word representations.

\subsubsection{2023—2024: Strengthening Connections to Real-World Scenarios}
Two past shared tasks are both characterized by explicit intention to put the theoretical problem closer to real-world scenarios and addressed the problem of possible topic consistency within the documents introducing controlled levels of thematic homogeneity in benchmark data challenging participants with development of methods less dependent on thematic signal. 

The solutions submitted to PAN $2023$ demonstrated relative difficulty of this setting. Whereas most submission achieved F1 score of more than $90$\% and $80$\% on \texttt{EASY} and \texttt{MEDIUM} tasks where writing style change could coincide with thematic shift, the performance on single-topic \texttt{HARD} dataset was significantly lower. 

The year was marked by further expansion of the PLMs' use, although one solution focusing on traditional stylometry was also submitted \cite{jacobo_authorship_2023}. One of the important tendencies that year was a broad diversity of ways in which PLMs's linguistic knowledge was integrated into the solutions. \cite{ye_supervised_2023, chen_contrastive_2023} made recourse to contrastive learning in former case combining it with a prompt-based approach that excelled on the \texttt{HARD} dataset. \cite{kucukkaya_arc-nlp_2023} solved the task as inference problem concatenating paragraphs and predicting special tokens \texttt{<ENTAILMENT>} or \texttt{<CONTRADICTION>}. \cite{huang_encoded_2023} pre-trained a custom model serving as a basis for classifier, while \cite{hashemi_enhancing_2023} achieved highest scores on \texttt{EASY} and \texttt{MEDIUM} data using an ensemble of several PLMs to predict binary labels for concatenated pairs of adjacent paragraphs.

The following year’s shared task on multi-author style analysis retained the same definition and structure of benchmark data as $2023$ \cite{zangerle_overview_2024}, continuing the focus on paragraph-level style change detection with varying levels of topical homogeneity. Contrastive learning techniques and ensemble architectures based on large PLMs took an even more prominent role. As a result, overall performance improved and the gap between multi-topic and single-topic document scenarios narrowed. The best submission achieved an impressive F1 around 86\% on \texttt{HARD} dataset. Notably, purely traditional stylometry approaches virtually disappeared in 2024, as nearly all top methods relied on fine-tuned transformer models (often in combination or with specialized training objectives) to detect writing style changes.

\subsubsection{Generated Text Detection in Human-AI Collaborative Hybrid Texts}

The surge of Generative AI has given rise to a new field of application for methods conceptually related to style change detection---authorship analysis in hybrid documents, i.e., texts co-written by humans and AI. \citeauthor{zeng_detecting_2024} investigated the detection of AI-generated passages within human–AI collaborative texts, highlighting unique challenges of this setting: frequent author switches, obfuscation by post-editing, and the limited availability of stylistic cues in short segments. The suggested approach---a two-step segmentation and classification method augmented with modern transformers and contrastive techniques---builds directly on the foundations laid by style change detection research.

This hybrid document segmentation problem was also the focus of the ALTA 2024 shared task \cite{molla_overview_2024}, which required participants to identify AI-generated sentences in news articles. The competition demonstrated a clear methodological convergence with style change detection research at PAN.

\section{Methodology}
\subsection{Challenges of Style Change Detection task at PAN 2025}

At first glance, this year's task may appear less challenging. On the one hand, it does not require the explicit grouping of text segments by authorship. On the other, sentence-level granularity is by no means  new and has been successfully addressed in previous editions. Yet, as the organizers note, the benchmark data was designed to more accurately replicate real-world scenarios, which is why the level of thematic coherence within each document was meticulously controlled \cite{bevendorff_overview_2025}.

Whereas the documents in the \texttt{EASY} dataset always cover multiple topics, the \texttt{MEDIUM} and \texttt{HARD} datasets exhibit limited or no thematic diversity, respectively. Therefore, while in the first case solutions may rely on thematic clues as potential indicators of style change, the latter two---each to a different extent---force participants to rely more heavily on detecting subtle style changes rather than topic variations \cite[438]{bevendorff_overview_2025}.

Further challenges of this year's shared task become evident in preliminary exploratory analysis of the benchmark data, which reveals several peculiarities of the data that significantly amplify the difficulty of the task (see Appendix \ref{app:data_stats}):  

\begin{itemize}
    \item Relatively short sentence length.
    \item A substantial portion---over 10\%---of all sentences are \emph{exact duplicates}, with some occurring more than  3{,}000 times\footnote{The most frequently repeated sentences include moderation messages (e.g., \textquote{Debate/discuss/argue the merits of ideas, don't attack people.}, \textquote{r/politics is currently accepting new moderator applications.}) and automatic notifications (e.g., \textquote{Personal insults, shill or troll accusations, hate speech, any suggestion or support of harm, violence, or death, and other rule violations can result in a permanent ban.}, \textquote{I am a bot, and this action was performed automatically.})}. 
    \item Hundreds of sequences only contain  punctuation marks, but are placed on separate lines and thus formally treated as separate sentences by the compilers of the data.
\end{itemize}

Consequently, each individual sentence may not provide sufficient information for identifying author's fingerprint and making a reliable decision. A pair of identical or nearly identical one- or two-word sentences---not only fairly common in Internet communication in general and also abundantly present in the data---may be entirely style- or even content-neutral, i.e., provide no reliable clues whatsoever.

\subsection{Core Intuition}
To address the notorious problem of such \textquote{shallow sentences}, we pivoted our approach around the idea of incorporating into the decision-making process the one thing that even the most minimalist one-word sentence always has---its context, or, in simpler terms, its position \emph{within the problem}. Therefore, we designed a BiLSTM-powered solution intended to model a problem \emph{as a whole} and capture positional dependencies between the document's sentences treated as atomic units that are organized into stylistically cohesive segments. 

Our inspiration comes from late $2010$s work on text segmentation and learning cohesion breaks. \cite{sehikh_topic_2017} demonstrated efficiency of bidirectional RNN trained on positive and negative examples of cohesive text in learning breaks in speech transcriptions. Further theoretical step was made by \cite{koshorek_text_2018} who presented text topic segmentation task as a supervised, specifically---sequence labeling, problem and employed to a BiLSTM powered architecture operating over sentence embeddings to implement this approach. \cite{li_segbot_2018}'s system, SegBot, achieved reliable performance on segmentation task at sentence and Elementary Discourse Unit (EDU) level. Improving and expanding the method, \cite{arnold_sector_2019} implemented a system segmenting texts into thematically coherent sections and assigning topic labels. \citeauthor{glavas_two-level_2020} and \citeauthor{lo_transformer_2021} implemented similar contextualization approaches employing two-level transformers. 

The approach also has predecessors among PAN participants. \citeauthor{hosseinia_parallel_2018} treated representations of problem's sentences with syntactic features as atomic units and explored their sequential dependencies using an LSTM. More recently, BiLSTM appeared several times as a steps in extraction of sentence or paragraph representation from PLMs \cite{deibel_style_2021, zia_style_2022}.

\subsection{Architecture}
\begin{figure}
  \centering  \includegraphics[width=\linewidth]{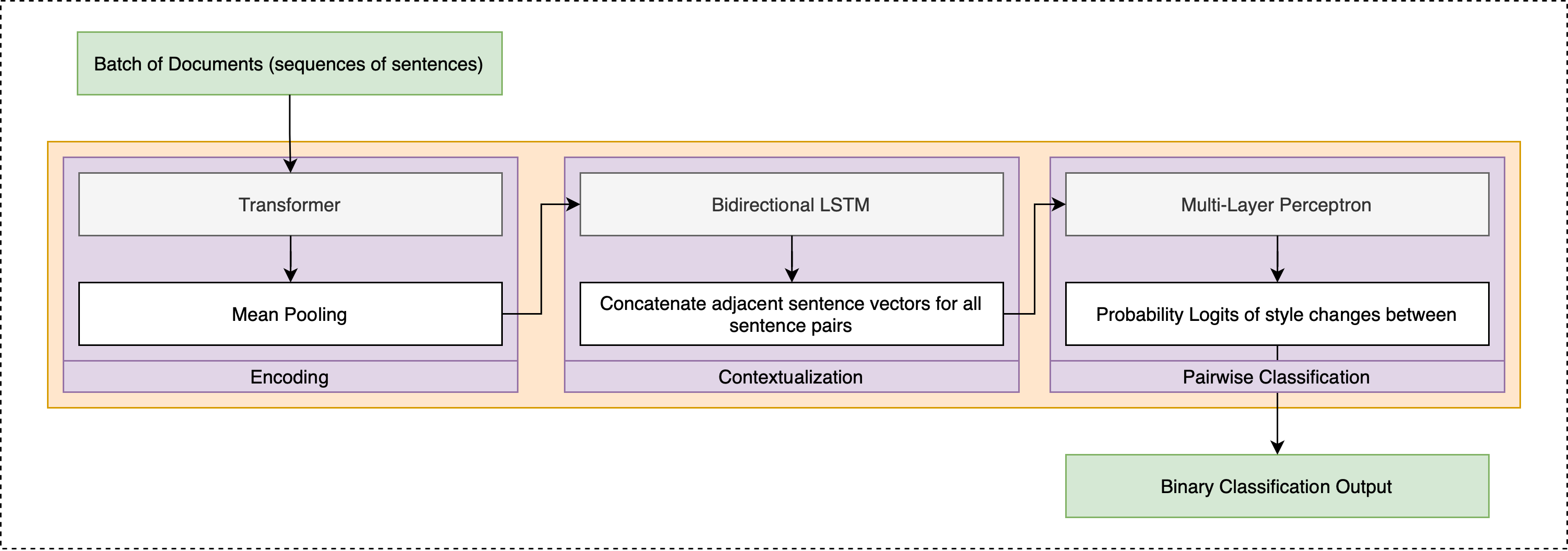}
  \caption{Sequential Sentence Pair Classifier}
  \label{fig:architecture}
\end{figure}

To implement the idea, we opted for a lightweight solution. A problem---a sequence of sentences---is considered a single sample, and its sentences are encoded using a PLM. Applying straightforward mean pooling to token embeddings, fixed-length representations of the problem's sentences are obtained. Then, to capture the inter-sentence contextual clues, the sequence of problem's sentence vectors is feed into a BiLSTM. This layer outputs context-aware sentence representations enriching raw mean-pooled vectors with information from the sentences across the entire problem. Subsequently, by concatenating each sentence vector with adjacent one across feature dimensions, representations of pairs of adjacent sentences are constructed. These are then passed through a multi-layer perceptron (MLP) classifier that outputs logits corresponding to the probability of a style change between each sentence pair (see Figure \ref{fig:architecture}).

This design enables the model to leverage both the semantic richness of the fine-tuned backbone PLM and the sequential structure of the problem, resulting in robust style change detection performance across documents of varying lengths and complexities. The submitted implementation code is available on GitHub\footnote{\url{https://github.com/glsch/better-call-claude_pan25-multi-author-style-analysis}}.

\subsubsection{Base Transformer}

Different pre-trained models were tested (see Table~\ref{tab:f1_scores}), but \texttt{StyleDistance/styledistance}, presented by \cite{patel_styledistance_2024}, was retained and submitted for evaluation on the test data. 

\begin{table}[ht]
\centering
\caption{F1 scores for each model, task, and dataset configuration.}
\label{tab:f1_scores}
\begin{tabular}{lccc|ccc}
\toprule
\textbf{Model} & \multicolumn{3}{c|}{\textbf{2025 data}} & \multicolumn{3}{c}{\textbf{2025 augmented with 2024 data}} \\
 & Easy & Medium & Hard & Easy & Medium & Hard \\
\midrule
RoBERTa-base  \cite{zhuang_robustly_2021}     & \textbf{0.929} & 0.764 & 0.613 & \textbf{0.930} & 0.787 & 0.656 \\
XLM-RoBERTa-base \cite{conneau_unsupervised_2020}  & 0.927 & 0.779 & 0.540 & 0.929 & 0.779 & 0.658 \\
all-MiniLM-L6-v2 \cite{reimers_sentence-bert_2019}  & 0.878 & 0.763 & 0.607 & 0.906 & 0.777 & 0.654 \\
StyleDistance \cite{patel_styledistance_2024}     & 0.891 & \textbf{0.780} & \textbf{0.629} & 0.922 & \textbf{0.828} & \textbf{0.723}\\
\bottomrule
\end{tabular}
\end{table}

\section{Training}
\subsection{Data Augmentation}
To obtain more training data, we used three datasets from PAN~$2024$. All sentence transitions within a single paragraph were labeled as not representing a style change. The first sentence of each new paragraph was labeled as $1$, indicating a style change.

A single model for all subtasks was trained on the most complete training data: the three sentence-level datasets from $2025$ and all three paragraph-turned-sentence-level datasets from $2024$.

\subsection{Training}
The Sequential Sentence Pair Classifier was implemented in PyTorch Lightning, allowing for seamless experimentation with both the architecture and hyperparameters.

\begin{table}[h]
\centering
\small
\caption{Model hyperparameters.}
\begin{tabular}{ll}
\toprule
\textbf{Hyperparameter} & \\
\midrule
Base transformer & \texttt{StyleDistance/styledistance} \\
Base transformer frozen & True \\
Pooling & Mean \\
BiLSTM layers & 5 \\
BiLSTM dropout & 0.2 \\
MLP & A three-layer feedforward network with linear layers, GELU activations, and dropout\\
MLP dropout & 0.2 \\
Batch & 4 \\
Learning rate & 0.0005 \\
Minimal learning rate & 0.00005 \\
Scheduler & cosine \\
Training steps & 30000 \\
Warmup steps & 2600 \\
Loss & Binary Cross-Entropy \\

\bottomrule
\end{tabular}
\label{tab:hparams}
\end{table}

\begin{table}[h]
\centering
\caption{Performance of Claude and Sequential Sentence Pair Classifier on the official PAN 2025 test data.}
\begin{tabular}{llr}
\toprule
Model & Dataset & F1 (macro) \\
\midrule
  & hard  & 0.661\\ 
Claude & medium  & \textbf{0.818} \\
 & easy  & 0.856 \\
 \midrule
 \multirow{3}{*}{Sequential Sentence Pair Classifier}
 & hard & \textbf{0.731}\\
 & medium   & 0.815 \\
 & easy & \textbf{0.929}  \\
\bottomrule
\end{tabular}
\label{tab:results_full}
\end{table}

\section{Baselines}
At PAN $2024$, three baselines were used: \texttt{RANDOM}, \texttt{PREDICT $1$}, and \texttt{PREDICT $0$}. While using all these baselines, we decided on another---and more challenging---one, \texttt{LLM}, zero-shot predictions by a best-performing LLM \texttt{claude-3.7-sonnet} prompted with the so-called linguistically informed prompts (LIP) \cite{huang_can_2024}. For the detailed description of the baseline setup we address the reader to \cite{schmidt_better_2025}.

\section{Results}

The following results were obtained on the official PAN 2025 test data (Table \ref{tab:results_full}). 

The gradual decline in performance from the easy to the hard tasks reflects the increasing difficulty of identifying style changes. Overall, our model outperforms zero-shot large language model predictions on both the easy and hard tasks and falls short by only a fraction of a percentage point on the medium task. At the same time, the proposed solution is lightweight and does not require much computational resources.

\section{Discussion}

Several observations stem from our approach to the PAN $2025$ style change detection task. 

First, while our model is effective, its strength lies in exploiting the macrostructure of the problem and inter-sentence contextual patterns---particularly the sequential order and distribution of sentences---rather than in isolating purely stylistic signals. This reflects a broader shift from traditional stylometric analysis, which typically assumes topical uniformity and relies on intrinsic features like syntax and lexical choice. This raises concerns about generalizability. 

Future iterations of the task could focus more on isolation of stylistic signal by reducing contextual cues---e.g., further controlling topic coherence or randomizing sentence order---to more rigorously test a model's ability to capture intrinsic authorial style.

Second, the strong zero-shot performance of Claude draws attention to the growing impact of LLMs in authorship analysis. LLMs, with their vast pretraining and generalization capabilities, can recognize both contextual and stylistic patterns with little to no task-specific adjustment. Future PAN tasks might clearly separate evaluation tracks that allow external LLM calls from those that do not even for otherwise AI-unrelated tasks.

\section*{Conclusion}

Despite the promising results of the proposed model, it has several limitations. First, the padding strategy required for batch processing may hinder scalability and efficiency when applied to much longer texts or inputs with highly variable lengths. Second, while the BiLSTM used for contextualization has proven effective, it may not be the optimal architecture for capturing complex dependencies, particularly in longer input sequences. More sophisticated architectures—such as those proposed by \cite{glavas_two-level_2020, lo_transformer_2021}—could potentially yield better results.

Finally, our reliance on a frozen pre-trained encoder may limit the model’s adaptability to domain-specific nuances. Fine-tuning the encoder or incorporating domain-specific data and training strategies could further improve performance.

\section*{Declaration on Generative AI}

 During the preparation of this work, the author(s) used Claude 4 and GPT-4 (gpt-3.5/gpt-4) to perform grammar/ spelling checks and edit text for clarity. Additionally, the author(s) used Perplexity’s research tools to double-check relevant literature and contributions related to the topic. All content was reviewed and edited by the author(s), who take full responsibility for the final publication.

\bibliography{sample-ceur, references}

\begin{thebibliography}{68}
\expandafter\ifx\csname natexlab\endcsname\relax\def\natexlab#1{#1}\fi
\providecommand{\url}[1]{\texttt{#1}}
\providecommand{\href}[2]{#2}
\providecommand{\path}[1]{#1}
\providecommand{\DOIprefix}{doi:}
\providecommand{\ArXivprefix}{arXiv:}
\providecommand{\URLprefix}{URL: }
\providecommand{\Pubmedprefix}{pmid:}
\providecommand{\doi}[1]{\href{http://dx.doi.org/#1}{\path{#1}}}
\providecommand{\Pubmed}[1]{\href{pmid:#1}{\path{#1}}}
\providecommand{\bibinfo}[2]{#2}
\ifx\xfnm\relax \def\xfnm[#1]{\unskip,\space#1}\fi
\bibitem[{Clérice and Glaise(2023)}]{clerice_twenty-one_2023}
\bibinfo{author}{T.~Clérice}, \bibinfo{author}{A.~Glaise},
\newblock \bibinfo{title}{Twenty-{One}* {Pseudo}-{Chrysostoms} and more: authorship verification in the patristic world},
\newblock in: \bibinfo{booktitle}{Computational {Humanities} {Research} {Conference} 2023}, Proceedings of the {Computational} {Humanities} {Research} {Conference} 2022, \bibinfo{address}{Paris, France}, \bibinfo{year}{2023}. \URLprefix \url{https://inria.hal.science/hal-04211176}.
\bibitem[{Cafiero and Camps(2021)}]{cafiero_psycheas_2021}
\bibinfo{author}{F.~Cafiero}, \bibinfo{author}{J.-B. Camps},
\newblock \bibinfo{title}{‘{Psyché}’as a {Rosetta} {Stone}? {Assessing} {Collaborative} {Authorship} in the {French} 17th {Century} {Theatre}},
\newblock \bibinfo{journal}{Proceedings http://ceur-ws. org ISSN} \bibinfo{volume}{1613} (\bibinfo{year}{2021}) \bibinfo{pages}{0073}.
\bibitem[{Plecháč(2021)}]{plechac_relative_2021}
\bibinfo{author}{P.~Plecháč},
\newblock \bibinfo{title}{Relative contributions of {Shakespeare} and {Fletcher} in {Henry} {VIII}: {An} analysis based on most frequent words and most frequent rhythmic patterns},
\newblock \bibinfo{journal}{Digital Scholarship in the Humanities} \bibinfo{volume}{36} (\bibinfo{year}{2021}) \bibinfo{pages}{430--438}. \bibinfo{note}{Publisher: Oxford University Press}.
\bibitem[{Schmidt et~al.(2024)Schmidt, Vybornaya, and Yamshchikov}]{schmidt_fine-tuning_2024}
\bibinfo{author}{G.~Schmidt}, \bibinfo{author}{V.~Vybornaya}, \bibinfo{author}{I.~P. Yamshchikov},
\newblock \bibinfo{title}{Fine-{Tuning} {Pre}-{Trained} {Language} {Models} for {Authorship} {Attribution} of the {Pseudo}-{Dionysian} {Ars} {Rhetorica}},
\newblock \bibinfo{address}{Aarhus}, \bibinfo{year}{2024}.
\bibitem[{Eder(2016)}]{eder_rolling_2016}
\bibinfo{author}{M.~Eder},
\newblock \bibinfo{title}{Rolling stylometry},
\newblock \bibinfo{journal}{Digital Scholarship in the Humanities} \bibinfo{volume}{31} (\bibinfo{year}{2016}) \bibinfo{pages}{457--469}. \bibinfo{note}{Publisher: Oxford University Press}.
\bibitem[{Bevendorff et~al.(2025)Bevendorff, Dementieva, Fröbe, Gipp, Greiner-Petter, Karlgren, Mayerl, Nakov, Panchenko, Potthast, Shelmanov, Stamatatos, Stein, Wang, Wiegmann, and Zangerle}]{bevendorff_overview_2025}
\bibinfo{author}{J.~Bevendorff}, \bibinfo{author}{D.~Dementieva}, \bibinfo{author}{M.~Fröbe}, \bibinfo{author}{B.~Gipp}, \bibinfo{author}{A.~Greiner-Petter}, \bibinfo{author}{J.~Karlgren}, \bibinfo{author}{M.~Mayerl}, \bibinfo{author}{P.~Nakov}, \bibinfo{author}{A.~Panchenko}, \bibinfo{author}{M.~Potthast}, \bibinfo{author}{A.~Shelmanov}, \bibinfo{author}{E.~Stamatatos}, \bibinfo{author}{B.~Stein}, \bibinfo{author}{Y.~Wang}, \bibinfo{author}{M.~Wiegmann}, \bibinfo{author}{E.~Zangerle},
\newblock \bibinfo{title}{Overview of {PAN} 2025: {Generative} {AI} {Detection}, {Multilingual} {Text} {Detoxification}, {Multi}-author {Writing} {Style} {Analysis}, and {Generative} {Plagiarism} {Detection}},
\newblock in: \bibinfo{editor}{C.~Hauff}, \bibinfo{editor}{C.~Macdonald}, \bibinfo{editor}{D.~Jannach}, \bibinfo{editor}{G.~Kazai}, \bibinfo{editor}{F.~M. Nardini}, \bibinfo{editor}{F.~Pinelli}, \bibinfo{editor}{F.~Silvestri}, \bibinfo{editor}{N.~Tonellotto} (Eds.), \bibinfo{booktitle}{Advances in {Information} {Retrieval}}, \bibinfo{publisher}{Springer Nature Switzerland}, \bibinfo{address}{Cham}, \bibinfo{year}{2025}, pp. \bibinfo{pages}{434--441}. \URLprefix \url{https://doi.org/10.1007/978-3-031-88720-8_64}. \DOIprefix\doi{10.1007/978-3-031-88720-8_64}.
\bibitem[{Zangerle et~al.(2025)Zangerle, Mayerl, Potthast, and Stein}]{zangerle_multi-author_2025}
\bibinfo{author}{E.~Zangerle}, \bibinfo{author}{M.~Mayerl}, \bibinfo{author}{M.~Potthast}, \bibinfo{author}{B.~Stein}, \bibinfo{title}{Multi-{Author} {Writing} {Style} {Analysis} 2025}, \bibinfo{year}{2025}. \URLprefix \url{https://pan.webis.de/clef25/pan25-web/style-change-detection.html}.
\bibitem[{Stamatatos et~al.(2016)Stamatatos, Tschnuggnall, Verhoeven, Daelemans, Specht, Stein, and Potthast}]{stamatatos_clustering_2016}
\bibinfo{author}{E.~Stamatatos}, \bibinfo{author}{M.~Tschnuggnall}, \bibinfo{author}{B.~Verhoeven}, \bibinfo{author}{W.~Daelemans}, \bibinfo{author}{G.~Specht}, \bibinfo{author}{B.~Stein}, \bibinfo{author}{M.~Potthast},
\newblock \bibinfo{title}{Clustering by authorship within and across documents},
\newblock in: \bibinfo{booktitle}{Working {Notes} {Papers} of the {CLEF} 2016 {Evaluation} {Labs}. {CEUR} {Workshop} {Proceedings}/{Balog}, {Krisztian} [edit.]; et al.}, \bibinfo{year}{2016}, pp. \bibinfo{pages}{691--715}.
\bibitem[{Rosso et~al.(2016)Rosso, Rangel, Potthast, Stamatatos, Tschuggnall, and Stein}]{rosso_overview_2016}
\bibinfo{author}{P.~Rosso}, \bibinfo{author}{F.~Rangel}, \bibinfo{author}{M.~Potthast}, \bibinfo{author}{E.~Stamatatos}, \bibinfo{author}{M.~Tschuggnall}, \bibinfo{author}{B.~Stein},
\newblock \bibinfo{title}{Overview of {PAN}’16: new challenges for authorship analysis: cross-genre profiling, clustering, diarization, and obfuscation},
\newblock in: \bibinfo{booktitle}{Experimental {IR} {Meets} {Multilinguality}, {Multimodality}, and {Interaction}: 7th {International} {Conference} of the {CLEF} {Association}, {CLEF} 2016, Évora, {Portugal}, {September} 5-8, 2016, {Proceedings} 7}, \bibinfo{publisher}{Springer}, \bibinfo{year}{2016}, pp. \bibinfo{pages}{332--350}.
\bibitem[{Kuznetsov et~al.(2016)Kuznetsov, Motrenko, Kuznetsova, and Strijov}]{kuznetsov_methods_2016}
\bibinfo{author}{M.~P. Kuznetsov}, \bibinfo{author}{A.~Motrenko}, \bibinfo{author}{R.~Kuznetsova}, \bibinfo{author}{V.~V. Strijov},
\newblock \bibinfo{title}{Methods for {Intrinsic} {Plagiarism} {Detection} and {Author} {Diarization}.},
\newblock in: \bibinfo{booktitle}{{CLEF} ({Working} notes)}, \bibinfo{year}{2016}, pp. \bibinfo{pages}{912--919}.
\bibitem[{Sittar et~al.(2016)Sittar, Iqbal, and Nawab}]{sittar_author_2016}
\bibinfo{author}{A.~Sittar}, \bibinfo{author}{H.~R. Iqbal}, \bibinfo{author}{R.~M.~A. Nawab},
\newblock \bibinfo{title}{Author {Diarization} {Using} {Cluster}-{Distance} {Approach}.},
\newblock in: \bibinfo{booktitle}{{CLEF} ({Working} {Notes})}, \bibinfo{year}{2016}, pp. \bibinfo{pages}{1000--1007}.
\bibitem[{Tschuggnall et~al.(2017)Tschuggnall, Stamatatos, Verhoeven, Daelemans, Specht, Stein, and Potthast}]{tschuggnall_overview_2017}
\bibinfo{author}{M.~Tschuggnall}, \bibinfo{author}{E.~Stamatatos}, \bibinfo{author}{B.~Verhoeven}, \bibinfo{author}{W.~Daelemans}, \bibinfo{author}{G.~Specht}, \bibinfo{author}{B.~Stein}, \bibinfo{author}{M.~Potthast},
\newblock \bibinfo{title}{Overview of the author identification task at {PAN}-2017: style breach detection and author clustering},
\newblock in: \bibinfo{booktitle}{Working {Notes} {Papers} of the {CLEF} 2017 {Evaluation} {Labs}/{Cappellato}, {Linda} [edit.]; et al.}, \bibinfo{year}{2017}, pp. \bibinfo{pages}{1--22}.
\bibitem[{Karas et~al.(2017)Karas, Spiewak, and Sobecki}]{karas_opi-jsa_2017}
\bibinfo{author}{D.~Karas}, \bibinfo{author}{M.~Spiewak}, \bibinfo{author}{P.~Sobecki},
\newblock \bibinfo{title}{{OPI}-{JSA} at {CLEF} 2017: {Author} {Clustering} and {Style} {Breach} {Detection}.},
\newblock in: \bibinfo{booktitle}{{CLEF} ({Working} {Notes})}, \bibinfo{year}{2017}.
\bibitem[{Khan(2017)}]{khan_style_2017}
\bibinfo{author}{J.~A. Khan},
\newblock \bibinfo{title}{Style {Breach} {Detection}: {An} {Unsupervised} {Detection} {Model}.},
\newblock in: \bibinfo{booktitle}{{CLEF} ({Working} {Notes})}, \bibinfo{year}{2017}.
\bibitem[{Safin and Kuznetsova(2017)}]{safin_style_2017}
\bibinfo{author}{K.~Safin}, \bibinfo{author}{R.~Kuznetsova},
\newblock \bibinfo{title}{Style {Breach} {Detection} with {Neural} {Sentence} {Embeddings}.},
\newblock in: \bibinfo{booktitle}{{CLEF} ({Working} {Notes})}, \bibinfo{year}{2017}.
\bibitem[{Kestemont et~al.(2018)Kestemont, Tschuggnall, Stamatatos, Daelemans, Specht, Stein, and Potthast}]{kestemont_overview_2018}
\bibinfo{author}{M.~Kestemont}, \bibinfo{author}{M.~Tschuggnall}, \bibinfo{author}{E.~Stamatatos}, \bibinfo{author}{W.~Daelemans}, \bibinfo{author}{G.~Specht}, \bibinfo{author}{B.~Stein}, \bibinfo{author}{M.~Potthast},
\newblock \bibinfo{title}{Overview of the author identification task at {PAN}-2018: cross-domain authorship attribution and style change detection},
\newblock in: \bibinfo{booktitle}{Working {Notes} {Papers} of the {CLEF} 2018 {Evaluation} {Labs}. {Avignon}, {France}, {September} 10-14, 2018/{Cappellato}, {Linda} [edit.]; et al.}, \bibinfo{year}{2018}, pp. \bibinfo{pages}{1--25}.
\bibitem[{Khan(2018)}]{khan_model_2018}
\bibinfo{author}{J.~A. Khan},
\newblock \bibinfo{title}{A model for style change detection at a glance},
\newblock volume \bibinfo{volume}{593}, \bibinfo{year}{2018}, p. \bibinfo{pages}{113}.
\bibitem[{Zlatkova et~al.(2018)Zlatkova, Kopev, Mitov, Atanasov, Hardalov, Koychev, and Nakov}]{zlatkova_ensemble-rich_2018}
\bibinfo{author}{D.~Zlatkova}, \bibinfo{author}{D.~Kopev}, \bibinfo{author}{K.~Mitov}, \bibinfo{author}{A.~Atanasov}, \bibinfo{author}{M.~Hardalov}, \bibinfo{author}{I.~Koychev}, \bibinfo{author}{P.~Nakov},
\newblock \bibinfo{title}{An ensemble-rich multi-aspect approach for robust style change detection},
\newblock in: \bibinfo{booktitle}{{CLEF} 2018 {Evaluation} {Labs} and {Workshop}–{Working} {Notes} {Papers}, {CEUR}-{WS}. org}, \bibinfo{year}{2018}, p.~\bibinfo{pages}{3}.
\bibitem[{Safin and Ogaltsov(2018)}]{safin_detecting_2018}
\bibinfo{author}{K.~Safin}, \bibinfo{author}{A.~Ogaltsov},
\newblock \bibinfo{title}{Detecting a change of style using text statistics},
\newblock \bibinfo{journal}{Working Notes of CLEF}  (\bibinfo{year}{2018}).
\bibitem[{Hosseinia and Mukherjee(2018)}]{hosseinia_parallel_2018}
\bibinfo{author}{M.~Hosseinia}, \bibinfo{author}{A.~Mukherjee},
\newblock \bibinfo{title}{A {Parallel} {Hierarchical} {Attention} {Network} for {Style} {Change} {Detection}},
\newblock \bibinfo{publisher}{CLEF}, \bibinfo{year}{2018}.
\bibitem[{Schaetti(2018)}]{schaetti_character-based_2018}
\bibinfo{author}{N.~Schaetti},
\newblock \bibinfo{title}{Character-based {Convolutional} {Neural} {Network} for {Style} {Change} {Detection}: {Notebook} for {PAN} at {CLEF} 2018.},
\newblock in: \bibinfo{booktitle}{{CLEF} ({Working} {Notes})}, \bibinfo{year}{2018}.
\bibitem[{Gorman(2020)}]{gorman_author_2020}
\bibinfo{author}{R.~Gorman},
\newblock \bibinfo{title}{Author identification of short texts using dependency treebanks without vocabulary},
\newblock \bibinfo{journal}{Digital Scholarship in the Humanities} \bibinfo{volume}{35} (\bibinfo{year}{2020}) \bibinfo{pages}{812--825}. \URLprefix \url{https://academic.oup.com/dsh/article/35/4/812/5606771}. \DOIprefix\doi{10.1093/llc/fqz070}.
\bibitem[{Gorman(2022)}]{gorman_universal_2022}
\bibinfo{author}{R.~Gorman},
\newblock \bibinfo{title}{Universal {Dependencies} and {Author} {Attribution} of {Short} {Texts} with {Syntax} {Alone}.},
\newblock \bibinfo{journal}{DHQ: Digital Humanities Quarterly} \bibinfo{volume}{16} (\bibinfo{year}{2022}).
\bibitem[{Gorman(2024)}]{gorman_morphosyntactic_2024}
\bibinfo{author}{R.~Gorman},
\newblock \bibinfo{title}{Morphosyntactic {Annotation} in {Literary} {Stylometry}},
\newblock \bibinfo{journal}{Information} \bibinfo{volume}{15} (\bibinfo{year}{2024}) \bibinfo{pages}{211}. \URLprefix \url{https://www.mdpi.com/2078-2489/15/4/211}. \DOIprefix\doi{10.3390/info15040211}.
\bibitem[{Gorman and Gorman(2024)}]{gorman_morphosyntactic_2024-1}
\bibinfo{author}{V.~B. Gorman}, \bibinfo{author}{R.~J. Gorman},
\newblock \bibinfo{title}{A morphosyntactic authorship attribution study of the speeches of {Demosthenes} and {Apollodorus}},
\newblock \bibinfo{journal}{The Journal of Hellenic Studies} \bibinfo{volume}{144} (\bibinfo{year}{2024}) \bibinfo{pages}{65--92}. \URLprefix \url{https://www.cambridge.org/core/product/identifier/S0075426924000302/type/journal_article}. \DOIprefix\doi{10.1017/S0075426924000302}.
\bibitem[{Zangerle et~al.(2019)Zangerle, Tschuggnall, Specht, Stein, and Potthast}]{zangerle_overview_2019}
\bibinfo{author}{E.~Zangerle}, \bibinfo{author}{M.~Tschuggnall}, \bibinfo{author}{G.~Specht}, \bibinfo{author}{B.~Stein}, \bibinfo{author}{M.~Potthast},
\newblock \bibinfo{title}{Overview of the {Style} {Change} {Detection} {Task} at {PAN} 2019},
\newblock in: \bibinfo{editor}{L.~Cappellato}, \bibinfo{editor}{N.~Ferro}, \bibinfo{editor}{D.~E. Losada}, \bibinfo{editor}{H.~Müller} (Eds.), \bibinfo{booktitle}{Working {Notes} {Papers} of the {CLEF} 2019 {Evaluation} {Labs}}, volume \bibinfo{volume}{2380} of \textit{\bibinfo{series}{{CEUR} {Workshop} {Proceedings}}}, \bibinfo{year}{2019}. \URLprefix \url{https://ceur-ws.org/Vol-2380/paper_243.pdf}, \bibinfo{note}{iSSN: 1613-0073}.
\bibitem[{Nath(2019)}]{nath_style_2019}
\bibinfo{author}{S.~Nath},
\newblock \bibinfo{title}{Style change detection by threshold based and window merge clustering methods.},
\newblock in: \bibinfo{booktitle}{{CLEF} ({Working} {Notes})}, \bibinfo{year}{2019}.
\bibitem[{Zuo et~al.(2019)Zuo, Zhao, and Banerjee}]{zuo_style_2019}
\bibinfo{author}{C.~Zuo}, \bibinfo{author}{Y.~Zhao}, \bibinfo{author}{R.~Banerjee},
\newblock \bibinfo{title}{Style {Change} {Detection} with {Feed}-forward {Neural} {Networks}.},
\newblock \bibinfo{journal}{CLEF (Working Notes)} \bibinfo{volume}{93} (\bibinfo{year}{2019}).
\bibitem[{Zangerle et~al.(2020)Zangerle, Mayerl, Specht, Stein, and Potthast}]{zangerle_overview_2020}
\bibinfo{author}{E.~Zangerle}, \bibinfo{author}{M.~Mayerl}, \bibinfo{author}{G.~Specht}, \bibinfo{author}{B.~Stein}, \bibinfo{author}{M.~Potthast},
\newblock \bibinfo{title}{Overview of the {Style} {Change} {Detection} {Task} at {PAN} 2020},
\newblock in: \bibinfo{editor}{L.~Cappellato}, \bibinfo{editor}{C.~Eickhoff}, \bibinfo{editor}{N.~Ferro}, \bibinfo{editor}{A.~Névéol} (Eds.), \bibinfo{booktitle}{Working {Notes} {Papers} of the {CLEF} 2020 {Evaluation} {Labs}}, volume \bibinfo{volume}{2696} of \textit{\bibinfo{series}{{CEUR} {Workshop} {Proceedings}}}, \bibinfo{year}{2020}. \URLprefix \url{https://ceur-ws.org/Vol-2696/paper_256.pdf}, \bibinfo{note}{iSSN: 1613-0073}.
\bibitem[{Iyer and Vosoughi(2020)}]{iyer_style_2020}
\bibinfo{author}{A.~Iyer}, \bibinfo{author}{S.~Vosoughi},
\newblock \bibinfo{title}{Style {Change} {Detection} {Using} {BERT}.},
\newblock \bibinfo{journal}{CLEF (Working Notes)} \bibinfo{volume}{93} (\bibinfo{year}{2020}) \bibinfo{pages}{106}.
\bibitem[{Castro-Castro et~al.(2020)Castro-Castro, Rodríguez-Lozada, and Muñoz}]{castro-castro_mixed_2020}
\bibinfo{author}{D.~Castro-Castro}, \bibinfo{author}{C.~A. Rodríguez-Lozada}, \bibinfo{author}{R.~Muñoz},
\newblock \bibinfo{title}{Mixed {Style} {Feature} {Representation} and {B}-maximal {Clustering} for {Style} {Change} {Detection}.},
\newblock in: \bibinfo{booktitle}{{CLEF} ({Working} {Notes})}, \bibinfo{year}{2020}.
\bibitem[{Zangerle et~al.(2021)Zangerle, Mayerl, Potthast, and Stein}]{zangerle_overview_2021}
\bibinfo{author}{E.~Zangerle}, \bibinfo{author}{M.~Mayerl}, \bibinfo{author}{M.~Potthast}, \bibinfo{author}{B.~Stein},
\newblock \bibinfo{title}{Overview of the {Style} {Change} {Detection} {Task} at {PAN} 2021},
\newblock in: \bibinfo{editor}{G.~Faggioli}, \bibinfo{editor}{N.~Ferro}, \bibinfo{editor}{A.~Joly}, \bibinfo{editor}{M.~Maistro}, \bibinfo{editor}{F.~Piroi} (Eds.), \bibinfo{booktitle}{Working {Notes} {Papers} of the {CLEF} 2021 {Evaluation} {Labs}}, volume \bibinfo{volume}{2936} of \textit{\bibinfo{series}{{CEUR} {Workshop} {Proceedings}}}, \bibinfo{year}{2021}. \URLprefix \url{https://ceur-ws.org/Vol-2936/paper-148.pdf}, \bibinfo{note}{iSSN: 1613-0073}.
\bibitem[{Zhang et~al.(2021)Zhang, Han, Kong, Miao, Peng, Zeng, Cao, Zhang, Xiao, and Peng}]{zhang_style_2021}
\bibinfo{author}{Z.~Zhang}, \bibinfo{author}{Z.~Han}, \bibinfo{author}{L.~Kong}, \bibinfo{author}{X.~Miao}, \bibinfo{author}{Z.~Peng}, \bibinfo{author}{J.~Zeng}, \bibinfo{author}{H.~Cao}, \bibinfo{author}{J.~Zhang}, \bibinfo{author}{Z.~Xiao}, \bibinfo{author}{X.~Peng},
\newblock \bibinfo{title}{Style {Change} {Detection} {Based} {On} {Writing} {Style} {Similarity}—{Notebook} for {PAN} at {CLEF} 2021},
\newblock in: \bibinfo{editor}{G.~Faggioli}, \bibinfo{editor}{N.~Ferro}, \bibinfo{editor}{A.~Joly}, \bibinfo{editor}{M.~Maistro}, \bibinfo{editor}{F.~Piroi} (Eds.), \bibinfo{booktitle}{{CLEF} 2021 {Labs} and {Workshops}, {Notebook} {Papers}}, \bibinfo{publisher}{CEUR-WS.org}, \bibinfo{year}{2021}. \URLprefix \url{http://ceur-ws.org/Vol-2936/paper-198.pdf}.
\bibitem[{Strøm(2021)}]{strom_multi-label_2021}
\bibinfo{author}{E.~Strøm},
\newblock \bibinfo{title}{Multi-label {Style} {Change} {Detection} by {Solving} a {Binary} {Classification} {Problem}.},
\newblock in: \bibinfo{booktitle}{{CLEF} (working notes)}, \bibinfo{year}{2021}, pp. \bibinfo{pages}{2146--2157}.
\bibitem[{Singh et~al.(2021)Singh, Weerasinghe, and Greenstadt}]{singh_writing_2021}
\bibinfo{author}{R.~Singh}, \bibinfo{author}{J.~Weerasinghe}, \bibinfo{author}{R.~Greenstadt},
\newblock \bibinfo{title}{Writing {Style} {Change} {Detection} on {Multi}-{Author} {Documents}.},
\newblock in: \bibinfo{booktitle}{{CLEF} ({Working} {Notes})}, \bibinfo{year}{2021}, pp. \bibinfo{pages}{2137--2145}.
\bibitem[{Weerasinghe and Greenstadt(2020)}]{weerasinghe_feature_2020}
\bibinfo{author}{J.~Weerasinghe}, \bibinfo{author}{R.~Greenstadt},
\newblock \bibinfo{title}{Feature vector difference based neural network and logistic regression models for authorship verification},
\newblock in: \bibinfo{booktitle}{{CEUR} workshop proceedings}, volume \bibinfo{volume}{2695}, \bibinfo{year}{2020}.
\bibitem[{Deibel and Löfflad(2021)}]{deibel_style_2021}
\bibinfo{author}{R.~Deibel}, \bibinfo{author}{D.~Löfflad},
\newblock \bibinfo{title}{Style {Change} {Detection} on {Real}-{World} {Data} using an {LSTM}-powered {Attribution} {Algorithm}.},
\newblock in: \bibinfo{booktitle}{{CLEF} ({Working} {Notes})}, \bibinfo{year}{2021}, pp. \bibinfo{pages}{1899--1909}.
\bibitem[{Nath(2021)}]{nath_style_2021}
\bibinfo{author}{S.~Nath},
\newblock \bibinfo{title}{Style change detection using {Siamese} neural networks.},
\newblock in: \bibinfo{booktitle}{{CLEF} ({Working} {Notes})}, \bibinfo{year}{2021}, pp. \bibinfo{pages}{2073--2082}.
\bibitem[{Alshamasi and Menai(2022)}]{alshamasi_ensemble-based_2022}
\bibinfo{author}{S.~Alshamasi}, \bibinfo{author}{M.~Menai},
\newblock \bibinfo{title}{Ensemble-{Based} {Clustering} for {Writing} {Style} {Change} {Detection} in {Multi}-{Authored} {Textual} {Documents}},
\newblock in: \bibinfo{editor}{G.~Faggioli}, \bibinfo{editor}{N.~Ferro}, \bibinfo{editor}{A.~Hanbury}, \bibinfo{editor}{M.~Potthast} (Eds.), \bibinfo{booktitle}{{CLEF} 2022 {Labs} and {Workshops}, {Notebook} {Papers}}, \bibinfo{publisher}{CEUR-WS.org}, \bibinfo{year}{2022}. \URLprefix \url{http://ceur-ws.org/Vol-3180/paper-187.pdf}.
\bibitem[{Alvi and Alqahtani(2022)}]{alvi_style_2022}
\bibinfo{author}{H.~A.~F. Alvi}, \bibinfo{author}{N.~Alqahtani},
\newblock \bibinfo{title}{Style {Change} {Detection} using {Discourse} {Markers}},
\newblock in: \bibinfo{editor}{G.~Faggioli}, \bibinfo{editor}{N.~Ferro}, \bibinfo{editor}{A.~Hanbury}, \bibinfo{editor}{M.~Potthast} (Eds.), \bibinfo{booktitle}{{CLEF} 2022 {Labs} and {Workshops}, {Notebook} {Papers}}, \bibinfo{publisher}{CEUR-WS.org}, \bibinfo{year}{2022}. \URLprefix \url{http://ceur-ws.org/Vol-3180/paper-188.pdf}.
\bibitem[{Rodríguez-Losada and Castro-Castro(2022)}]{rodriguez-losada_three_2022}
\bibinfo{author}{C.~A. Rodríguez-Losada}, \bibinfo{author}{D.~Castro-Castro},
\newblock \bibinfo{title}{Three {Style} {Similarity}: sentence-embedding, auxiliary words, punctuation},
\newblock in: \bibinfo{editor}{G.~Faggioli}, \bibinfo{editor}{N.~Ferro}, \bibinfo{editor}{A.~Hanbury}, \bibinfo{editor}{M.~Potthast} (Eds.), \bibinfo{booktitle}{{CLEF} 2022 {Labs} and {Workshops}, {Notebook} {Papers}}, \bibinfo{publisher}{CEUR-WS.org}, \bibinfo{year}{2022}. \URLprefix \url{http://ceur-ws.org/Vol-3180/paper-218.pdf}.
\bibitem[{Graner and Ranly(2022)}]{graner_unorthodox_2022}
\bibinfo{author}{L.~Graner}, \bibinfo{author}{P.~Ranly},
\newblock \bibinfo{title}{An {Unorthodox} {Approach} for {Style} {Change} {Detection}.},
\newblock in: \bibinfo{booktitle}{{CLEF} ({Working} {Notes})}, \bibinfo{year}{2022}, pp. \bibinfo{pages}{2455--2466}.
\bibitem[{Lin et~al.(2022)Lin, Chen, Tzeng, and Lee}]{lin_ensemble_2022}
\bibinfo{author}{T.-M. Lin}, \bibinfo{author}{C.-Y. Chen}, \bibinfo{author}{Y.-W. Tzeng}, \bibinfo{author}{L.-H. Lee},
\newblock \bibinfo{title}{Ensemble {Pre}-trained {Transformer} {Models} for {Writing} {Style} {Change} {Detection}.},
\newblock in: \bibinfo{booktitle}{{CLEF} ({Working} {Notes})}, \bibinfo{year}{2022}, pp. \bibinfo{pages}{2565--2573}.
\bibitem[{Lao et~al.(2022)Lao, Ma, Yang, Yang, Yuan, Tan, and Liang}]{lao_style_2022}
\bibinfo{author}{Q.~Lao}, \bibinfo{author}{L.~Ma}, \bibinfo{author}{W.~Yang}, \bibinfo{author}{Z.~Yang}, \bibinfo{author}{D.~Yuan}, \bibinfo{author}{Z.~Tan}, \bibinfo{author}{L.~Liang},
\newblock \bibinfo{title}{Style {Change} {Detection} {Based} {On} {Bert} {And} {Conv1d}},
\newblock in: \bibinfo{editor}{G.~Faggioli}, \bibinfo{editor}{N.~Ferro}, \bibinfo{editor}{A.~Hanbury}, \bibinfo{editor}{M.~Potthast} (Eds.), \bibinfo{booktitle}{{CLEF} 2022 {Labs} and {Workshops}, {Notebook} {Papers}}, \bibinfo{publisher}{CEUR-WS.org}, \bibinfo{year}{2022}. \URLprefix \url{http://ceur-ws.org/Vol-3180/paper-208.pdf}.
\bibitem[{Zhang et~al.(2022)Zhang, Han, and Kong}]{zhang_style_2022}
\bibinfo{author}{Z.~Zhang}, \bibinfo{author}{Z.~Han}, \bibinfo{author}{L.~Kong},
\newblock \bibinfo{title}{Style {Change} {Detection} based on {Prompt}.},
\newblock in: \bibinfo{booktitle}{{CLEF} ({Working} {Notes})}, \bibinfo{year}{2022}, pp. \bibinfo{pages}{2753--2756}.
\bibitem[{Jiang and Huang(2022)}]{jiang_style_2022}
\bibinfo{author}{Z.~Z.~X. Jiang}, \bibinfo{author}{M.~Huang},
\newblock \bibinfo{title}{Style {Change} {Detection}: {Method} {Based} {On} {Pre}-trained {Model} {And} {Similarity} {Recognition}},
\newblock in: \bibinfo{editor}{G.~Faggioli}, \bibinfo{editor}{N.~Ferro}, \bibinfo{editor}{A.~Hanbury}, \bibinfo{editor}{M.~Potthast} (Eds.), \bibinfo{booktitle}{{CLEF} 2022 {Labs} and {Workshops}, {Notebook} {Papers}}, \bibinfo{publisher}{CEUR-WS.org}, \bibinfo{year}{2022}. \URLprefix \url{http://ceur-ws.org/Vol-3180/paper-205.pdf}.
\bibitem[{Zia and Liua(2022)}]{zia_style_2022}
\bibinfo{author}{L.~Z.~J. Zia}, \bibinfo{author}{Z.~Liua},
\newblock \bibinfo{title}{Style {Change} {Detection} {Based} {On} {Bi}-{LSTM} {And} {Bert}},
\newblock in: \bibinfo{editor}{G.~Faggioli}, \bibinfo{editor}{N.~Ferro}, \bibinfo{editor}{A.~Hanbury}, \bibinfo{editor}{M.~Potthast} (Eds.), \bibinfo{booktitle}{{CLEF} 2022 {Labs} and {Workshops}, {Notebook} {Papers}}, \bibinfo{publisher}{CEUR-WS.org}, \bibinfo{year}{2022}. \URLprefix \url{http://ceur-ws.org/Vol-3180/paper-234.pdf}.
\bibitem[{Jacobo et~al.(2023)Jacobo, Dehesa, Rojas, and Gómez-Adorno}]{jacobo_authorship_2023}
\bibinfo{author}{G.~Jacobo}, \bibinfo{author}{V.~Dehesa}, \bibinfo{author}{D.~Rojas}, \bibinfo{author}{H.~Gómez-Adorno},
\newblock \bibinfo{title}{Authorship verification machine learning methods for {Style} {Change} {Detection} in texts},
\newblock in: \bibinfo{editor}{M.~Aliannejadi}, \bibinfo{editor}{G.~Faggioli}, \bibinfo{editor}{N.~Ferro}, \bibinfo{editor}{M.~Vlachos} (Eds.), \bibinfo{booktitle}{Working {Notes} of {CLEF} 2023 - {Conference} and {Labs} of the {Evaluation} {Forum}}, \bibinfo{publisher}{CEUR-WS.org}, \bibinfo{year}{2023}, pp. \bibinfo{pages}{2652--2658}. \URLprefix \url{https://ceur-ws.org/Vol-3497/paper-217.pdf}.
\bibitem[{Ye et~al.(2023)Ye, Zhong, Qi, and Han}]{ye_supervised_2023}
\bibinfo{author}{Z.~Ye}, \bibinfo{author}{C.~Zhong}, \bibinfo{author}{H.~Qi}, \bibinfo{author}{Y.~Han},
\newblock \bibinfo{title}{Supervised {Contrastive} {Learning} for {Multi}-{Author} {Writing} {Style} {Analysis}.},
\newblock in: \bibinfo{booktitle}{{CLEF} ({Working} {Notes})}, \bibinfo{year}{2023}, pp. \bibinfo{pages}{2817--2822}.
\bibitem[{Chen and Liu(2023)}]{chen_contrastive_2023}
\bibinfo{author}{W.~Chen}, \bibinfo{author}{X.~Liu},
\newblock \bibinfo{title}{Contrastive {Learning} {Approaches} for {Multi}-{Author} {Style} {Analysis}},
\newblock in: \bibinfo{booktitle}{{CLEF} 2023 {Working} {Notes}}, volume \bibinfo{volume}{3497}, \bibinfo{publisher}{CEUR-WS}, \bibinfo{year}{2023}.
\bibitem[{Kucukkaya et~al.(2023)Kucukkaya, Sahin, and Toraman}]{kucukkaya_arc-nlp_2023}
\bibinfo{author}{I.~E. Kucukkaya}, \bibinfo{author}{U.~Sahin}, \bibinfo{author}{C.~Toraman},
\newblock \bibinfo{title}{{ARC}-{NLP} at {PAN} 23: {Transition}-{Focused} {Natural} {Language} {Inference} for {Writing} {Style} {Detection}},
\newblock in: \bibinfo{editor}{M.~Aliannejadi}, \bibinfo{editor}{G.~Faggioli}, \bibinfo{editor}{N.~Ferro}, \bibinfo{editor}{M.~Vlachos} (Eds.), \bibinfo{booktitle}{Working {Notes} of {CLEF} 2023 - {Conference} and {Labs} of the {Evaluation} {Forum}}, \bibinfo{publisher}{CEUR-WS.org}, \bibinfo{year}{2023}, pp. \bibinfo{pages}{2659--2668}. \URLprefix \url{https://ceur-ws.org/Vol-3497/paper-218.pdf}.
\bibitem[{Huang et~al.(2023)Huang, Huang, and Kong}]{huang_encoded_2023}
\bibinfo{author}{M.~Huang}, \bibinfo{author}{Z.~Huang}, \bibinfo{author}{L.~Kong},
\newblock \bibinfo{title}{Encoded {Classifier} {Using} {Knowledge} {Distillation} for {Multi}-{Author} {Writing} {Style} {Analysis}},
\newblock in: \bibinfo{editor}{M.~Aliannejadi}, \bibinfo{editor}{G.~Faggioli}, \bibinfo{editor}{N.~Ferro}, \bibinfo{editor}{M.~Vlachos} (Eds.), \bibinfo{booktitle}{Working {Notes} of {CLEF} 2023 - {Conference} and {Labs} of the {Evaluation} {Forum}}, \bibinfo{publisher}{CEUR-WS.org}, \bibinfo{year}{2023}, pp. \bibinfo{pages}{2629--2634}. \URLprefix \url{https://ceur-ws.org/Vol-3497/paper-214.pdf}.
\bibitem[{Hashemi and Shi(2023)}]{hashemi_enhancing_2023}
\bibinfo{author}{A.~Hashemi}, \bibinfo{author}{W.~Shi},
\newblock \bibinfo{title}{Enhancing {Writing} {Style} {Change} {Detection} using {Transformer}-based {Models} and {Data} {Augmentation}.},
\newblock in: \bibinfo{booktitle}{{CLEF} ({Working} {Notes})}, \bibinfo{year}{2023}, pp. \bibinfo{pages}{2613--2621}.
\bibitem[{Zangerle et~al.(2024)Zangerle, Mayerl, Potthast, and Stein}]{zangerle_overview_2024}
\bibinfo{author}{E.~Zangerle}, \bibinfo{author}{M.~Mayerl}, \bibinfo{author}{M.~Potthast}, \bibinfo{author}{B.~Stein},
\newblock \bibinfo{title}{Overview of the {Multi}-{Author} {Writing} {Style} {Analysis} {Task} at {PAN} 2024},
\newblock in: \bibinfo{editor}{G.~Faggioli}, \bibinfo{editor}{N.~Ferro}, \bibinfo{editor}{P.~Galuščáková}, \bibinfo{editor}{A.~G.~S. Herrera} (Eds.), \bibinfo{booktitle}{Working {Notes} {Papers} of the {CLEF} 2024 {Evaluation} {Labs}}, \bibinfo{publisher}{CEUR-WS.org}, \bibinfo{year}{2024}, pp. \bibinfo{pages}{2513--2522}. \URLprefix \url{http://ceur-ws.org/Vol-3740/paper-222.pdf}.
\bibitem[{Zeng et~al.(2024)Zeng, Liu, Sha, Li, Yang, Liu, Gašević, and Chen}]{zeng_detecting_2024}
\bibinfo{author}{Z.~Zeng}, \bibinfo{author}{S.~Liu}, \bibinfo{author}{L.~Sha}, \bibinfo{author}{Z.~Li}, \bibinfo{author}{K.~Yang}, \bibinfo{author}{S.~Liu}, \bibinfo{author}{D.~Gašević}, \bibinfo{author}{G.~Chen}, \bibinfo{title}{Detecting {AI}-{Generated} {Sentences} in {Human}-{AI} {Collaborative} {Hybrid} {Texts}: {Challenges}, {Strategies}, and {Insights}}, \bibinfo{year}{2024}. \URLprefix \url{http://arxiv.org/abs/2403.03506}. \DOIprefix\doi{10.48550/arXiv.2403.03506}, \bibinfo{note}{arXiv:2403.03506 [cs]}.
\bibitem[{Mollá et~al.(2024)Mollá, Xu, Zeng, and Li}]{molla_overview_2024}
\bibinfo{author}{D.~Mollá}, \bibinfo{author}{Q.~Xu}, \bibinfo{author}{Z.~Zeng}, \bibinfo{author}{Z.~Li},
\newblock \bibinfo{title}{Overview of the 2024 alta shared task: {Detect} automatic ai-generated sentences for human-ai hybrid articles},
\newblock \bibinfo{journal}{arXiv preprint arXiv:2412.17848}  (\bibinfo{year}{2024}).
\bibitem[{Sehikh et~al.(2017)Sehikh, Fohr, and Illina}]{sehikh_topic_2017}
\bibinfo{author}{I.~Sehikh}, \bibinfo{author}{D.~Fohr}, \bibinfo{author}{I.~Illina},
\newblock \bibinfo{title}{Topic segmentation in {ASR} transcripts using bidirectional {RNNs} for change detection},
\newblock in: \bibinfo{booktitle}{2017 {IEEE} automatic speech recognition and understanding workshop ({ASRU})}, \bibinfo{publisher}{IEEE}, \bibinfo{year}{2017}, pp. \bibinfo{pages}{512--518}.
\bibitem[{Koshorek et~al.(2018)Koshorek, Cohen, Mor, Rotman, and Berant}]{koshorek_text_2018}
\bibinfo{author}{O.~Koshorek}, \bibinfo{author}{A.~Cohen}, \bibinfo{author}{N.~Mor}, \bibinfo{author}{M.~Rotman}, \bibinfo{author}{J.~Berant},
\newblock \bibinfo{title}{Text {Segmentation} as a {Supervised} {Learning} {Task}},
\newblock in: \bibinfo{booktitle}{Proceedings of the 2018 {Conference} of the {North} {American} {Chapter} of the {Association} for {Computational} {Linguistics}: {Human} {Language} {Technologies}, {Volume} 2 ({Short} {Papers})}, \bibinfo{publisher}{Association for Computational Linguistics}, \bibinfo{address}{New Orleans, Louisiana}, \bibinfo{year}{2018}, pp. \bibinfo{pages}{469--473}. \URLprefix \url{http://aclweb.org/anthology/N18-2075}. \DOIprefix\doi{10.18653/v1/N18-2075}.
\bibitem[{Li et~al.(2018)Li, Sun, and Joty}]{li_segbot_2018}
\bibinfo{author}{J.~Li}, \bibinfo{author}{A.~Sun}, \bibinfo{author}{S.~R. Joty},
\newblock \bibinfo{title}{{SegBot}: {A} {Generic} {Neural} {Text} {Segmentation} {Model} with {Pointer} {Network}.},
\newblock in: \bibinfo{booktitle}{{IJCAI}}, \bibinfo{year}{2018}, pp. \bibinfo{pages}{4166--4172}.
\bibitem[{Arnold et~al.(2019)Arnold, Schneider, Cudré-Mauroux, Gers, and Löser}]{arnold_sector_2019}
\bibinfo{author}{S.~Arnold}, \bibinfo{author}{R.~Schneider}, \bibinfo{author}{P.~Cudré-Mauroux}, \bibinfo{author}{F.~A. Gers}, \bibinfo{author}{A.~Löser},
\newblock \bibinfo{title}{{SECTOR}: {A} {Neural} {Model} for {Coherent} {Topic} {Segmentation} and {Classification}},
\newblock \bibinfo{journal}{Transactions of the Association for Computational Linguistics} \bibinfo{volume}{7} (\bibinfo{year}{2019}) \bibinfo{pages}{169--184}. \URLprefix \url{https://direct.mit.edu/tacl/article/43514}. \DOIprefix\doi{10.1162/tacl_a_00261}.
\bibitem[{Glavaš and Somasundaran(2020)}]{glavas_two-level_2020}
\bibinfo{author}{G.~Glavaš}, \bibinfo{author}{S.~Somasundaran},
\newblock \bibinfo{title}{Two-{Level} {Transformer} and {Auxiliary} {Coherence} {Modeling} for {Improved} {Text} {Segmentation}},
\newblock \bibinfo{journal}{Proceedings of the AAAI Conference on Artificial Intelligence} \bibinfo{volume}{34} (\bibinfo{year}{2020}) \bibinfo{pages}{7797--7804}. \URLprefix \url{https://ojs.aaai.org/index.php/AAAI/article/view/6284}. \DOIprefix\doi{10.1609/aaai.v34i05.6284}.
\bibitem[{Lo et~al.(2021)Lo, Jin, Tan, Liu, Du, and Buntine}]{lo_transformer_2021}
\bibinfo{author}{K.~Lo}, \bibinfo{author}{Y.~Jin}, \bibinfo{author}{W.~Tan}, \bibinfo{author}{M.~Liu}, \bibinfo{author}{L.~Du}, \bibinfo{author}{W.~Buntine}, \bibinfo{title}{Transformer over {Pre}-trained {Transformer} for {Neural} {Text} {Segmentation} with {Enhanced} {Topic} {Coherence}}, \bibinfo{year}{2021}. \URLprefix \url{http://arxiv.org/abs/2110.07160}. \DOIprefix\doi{10.48550/arXiv.2110.07160}, \bibinfo{note}{arXiv:2110.07160 [cs]}.
\bibitem[{Patel et~al.(2024)Patel, Zhu, Qiu, Horvitz, Apidianaki, McKeown, and Callison-Burch}]{patel_styledistance_2024}
\bibinfo{author}{A.~Patel}, \bibinfo{author}{J.~Zhu}, \bibinfo{author}{J.~Qiu}, \bibinfo{author}{Z.~Horvitz}, \bibinfo{author}{M.~Apidianaki}, \bibinfo{author}{K.~McKeown}, \bibinfo{author}{C.~Callison-Burch}, \bibinfo{title}{{StyleDistance}: {Stronger} {Content}-{Independent} {Style} {Embeddings} with {Synthetic} {Parallel} {Examples}}, \bibinfo{year}{2024}. \URLprefix \url{https://arxiv.org/abs/2410.12757}. \DOIprefix\doi{10.48550/ARXIV.2410.12757}, \bibinfo{note}{version Number: 2}.
\bibitem[{Zhuang et~al.(2021)Zhuang, Wayne, Ya, and Jun}]{zhuang_robustly_2021}
\bibinfo{author}{L.~Zhuang}, \bibinfo{author}{L.~Wayne}, \bibinfo{author}{S.~Ya}, \bibinfo{author}{Z.~Jun},
\newblock \bibinfo{title}{A {Robustly} {Optimized} {BERT} {Pre}-training {Approach} with {Post}-training},
\newblock in: \bibinfo{editor}{S.~Li}, \bibinfo{editor}{M.~Sun}, \bibinfo{editor}{Y.~Liu}, \bibinfo{editor}{H.~Wu}, \bibinfo{editor}{K.~Liu}, \bibinfo{editor}{W.~Che}, \bibinfo{editor}{S.~He}, \bibinfo{editor}{G.~Rao} (Eds.), \bibinfo{booktitle}{Proceedings of the 20th {Chinese} {National} {Conference} on {Computational} {Linguistics}}, \bibinfo{year}{2021}, pp. \bibinfo{pages}{1218--1227}. \URLprefix \url{https://aclanthology.org/2021.ccl-1.108/}.
\bibitem[{Conneau et~al.(2020)Conneau, Khandelwal, Goyal, Chaudhary, Wenzek, Guzmán, Grave, Ott, Zettlemoyer, and Stoyanov}]{conneau_unsupervised_2020}
\bibinfo{author}{A.~Conneau}, \bibinfo{author}{K.~Khandelwal}, \bibinfo{author}{N.~Goyal}, \bibinfo{author}{V.~Chaudhary}, \bibinfo{author}{G.~Wenzek}, \bibinfo{author}{F.~Guzmán}, \bibinfo{author}{E.~Grave}, \bibinfo{author}{M.~Ott}, \bibinfo{author}{L.~Zettlemoyer}, \bibinfo{author}{V.~Stoyanov},
\newblock \bibinfo{title}{Unsupervised {Cross}-lingual {Representation} {Learning} at {Scale}},
\newblock in: \bibinfo{editor}{D.~Jurafsky}, \bibinfo{editor}{J.~Chai}, \bibinfo{editor}{N.~Schluter}, \bibinfo{editor}{J.~Tetreault} (Eds.), \bibinfo{booktitle}{Proceedings of the 58th {Annual} {Meeting} of the {Association} for {Computational} {Linguistics}}, \bibinfo{year}{2020}, pp. \bibinfo{pages}{8440--8451}. \URLprefix \url{https://aclanthology.org/2020.acl-main.747/}. \DOIprefix\doi{10.18653/v1/2020.acl-main.747}.
\bibitem[{Reimers and Gurevych(2019)}]{reimers_sentence-bert_2019}
\bibinfo{author}{N.~Reimers}, \bibinfo{author}{I.~Gurevych},
\newblock \bibinfo{title}{Sentence-{BERT}: {Sentence} {Embeddings} using {Siamese} {BERT}-{Networks}},
\newblock in: \bibinfo{booktitle}{Proceedings of the 2019 {Conference} on {Empirical} {Methods} in {Natural} {Language} {Processing}}, \bibinfo{publisher}{Association for Computational Linguistics}, \bibinfo{year}{2019}. \URLprefix \url{https://arxiv.org/abs/1908.10084}.
\bibitem[{Huang et~al.(2024)Huang, Chen, and Shu}]{huang_can_2024}
\bibinfo{author}{B.~Huang}, \bibinfo{author}{C.~Chen}, \bibinfo{author}{K.~Shu}, \bibinfo{title}{Can {Large} {Language} {Models} {Identify} {Authorship}?}, \bibinfo{year}{2024}. \URLprefix \url{http://arxiv.org/abs/2403.08213}. \DOIprefix\doi{10.48550/arXiv.2403.08213}, \bibinfo{note}{arXiv:2403.08213 [cs]}.
\bibitem[{Schmidt et~al.(2025)Schmidt, Römisch, Yamshchikov, Gorovaia, and Halchynska}]{schmidt_better_2025}
\bibinfo{author}{G.~Schmidt}, \bibinfo{author}{J.~Römisch}, \bibinfo{author}{I.~Yamshchikov}, \bibinfo{author}{S.~Gorovaia}, \bibinfo{author}{M.~Halchynska},
\newblock \bibinfo{title}{Better {Call} {Claude}: {Can} {LLMs} {Detect} {Changes} of {Writing} {Style}?},
\newblock in: \bibinfo{booktitle}{Experimental {IR} {Meets} {Multilinguality}, {Multimodality}, and {Interaction}. {Proceedings} of the {Sixteenth} {International} {Conference} of the {CLEF} {Association} ({CLEF} 2025)}, \bibinfo{year}{2025}.

\end{thebibliography}

\appendix
\section{Dataset Statistics}
\label{app:data_stats}
Tables~\ref{tab:document_statistics} and \ref{tab:top5sents} represent general data statistics and top-5 duplicated sentences in the data.



\begin{table}[ht]
\footnotesize
\centering
\caption{Document statistics for different subsets of the dataset}

\begin{tabular}{lcccccc}
\toprule
\textbf{Subset} & \textbf{Problems} & \textbf{Sentences} & \textbf{Avg words/sent.} & \textbf{Median words/sent.} & \textbf{Avg sent./doc} & \textbf{Median sent./doc} \\
\midrule
Easy & 5100 & 63584 & 16.9 $\pm$ 16.4 & 13.0 & 12.5 $\pm$ 4.1 & 12.0 \\
Medium & 5100 & 76379 & 17.7 $\pm$ 11.9 & 15.0 & 15.0 $\pm$ 9.4 & 12.0 \\
Hard & 5100 & 66555 & 18.7 $\pm$ 12.0 & 17.0 & 13.0 $\pm$ 4.7 & 12.0 \\
All Data & 15300 & 206518 & 17.8 $\pm$ 13.5 & 15.0 & 13.5 $\pm$ 6.6 & 12.0 \\
\bottomrule
\end{tabular}

\label{tab:document_statistics}
\end{table}

\begin{table*}[h]
\centering
\small
\caption{Top-5 most frequent sentences.}
\begin{tabular}{p{0.85\textwidth}r}
\toprule
\textbf{Sentence} & \textbf{Count} \\
\midrule
In general, be courteous to others. & 3296 \\
Debate/discuss/argue the merits of ideas, don't attack people. & 3296 \\
Personal insults, shill or troll accusations, hate speech, any suggestion or support of harm, violence, or death, and other rule violations can result in a permanent ban. & 3296 \\
For those who have questions regarding any media outlets being posted on this subreddit, please click to review our details as to our approved domains list and outlet criteria. & 2953 \\
r/politics is currently accepting new moderator applications. & 2172 \\
\bottomrule
\end{tabular}
\label{tab:top5sents}
\end{table*}

\section{Online Resources}
\begin{itemize}
\item \href{https://github.com/glsch/better-call-claude_pan25-multi-author-style-analysis/}{GitHub}
\end{itemize}

\end{document}